\documentclass[10pt,twocolumn,letterpaper]{article}
\pdfoutput=1

\usepackage{cvpr}
\usepackage{times}
\usepackage{epsfig}
\usepackage{graphicx}
\usepackage{amsmath}
\usepackage{amssymb}
\usepackage{kotex}
\usepackage{enumitem}
\usepackage{verbatim}
\usepackage{amsthm}
\usepackage{bm}
\usepackage{booktabs,makecell,multirow, caption, tabularx}
\usepackage{algorithm}
\usepackage{algorithmic}
\usepackage{url}
\usepackage{balance}
\usepackage{upquote}
\usepackage{appendix}
\usepackage{nicefrac}
\usepackage{physics}
\usepackage{hyperref}
\hypersetup{pagebackref=true,breaklinks=true,letterpaper=true,colorlinks,bookmarks=false}

\cvprfinalcopy 

\def\cvprPaperID{7089} 

\ifcvprfinal\fi
\begin{document}
\title{Adversarial Vertex Mixup: Toward Better Adversarially Robust Generalization}

\author{Saehyung Lee~~~~~~~ Hyungyu Lee~~~~~~~ Sungroh Yoon\thanks{Correspondence to: Sungroh Yoon \href{mailto:sryoon@snu.ac.kr}{sryoon@snu.ac.kr}.}\\
Electrical and Computer Engineering, ASRI, INMC, and Institute of Engineering Research\\ Seoul National University, Seoul 08826, South Korea\\
{\tt\small \{halo8218, rucy74, sryoon\}@snu.ac.kr}}

\maketitle

\begin{abstract}
   Adversarial examples cause neural networks to produce incorrect outputs with high confidence. Although adversarial training is one of the most effective forms of defense against adversarial examples, unfortunately, a large gap exists between test accuracy and training accuracy in adversarial training. In this paper, 
   we identify Adversarial Feature Overfitting~(AFO), which may cause poor adversarially robust generalization, and we show that adversarial training can overshoot the optimal point in terms of robust generalization, leading to AFO in our simple Gaussian model. Considering these theoretical results, 
   we present soft labeling as a solution to the AFO problem.
   Furthermore, we propose Adversarial Vertex mixup~(AVmixup), a soft-labeled data augmentation approach for improving adversarially robust generalization. We complement our theoretical analysis with experiments on CIFAR10, CIFAR100, SVHN, and Tiny ImageNet, and show that AVmixup significantly improves the robust generalization performance and that it reduces the trade-off between standard accuracy and adversarial robustness.
\end{abstract}

\section{Introduction}


Deep neural networks~(DNNs) have produced impressive results for various machine learning tasks, including computer vision~\cite{imagenet_2012} and natural language processing~\cite{nlp_2012}. Neural networks, however, can be easily fooled by small adversarial perturbations of their input with a high degree of confidence~\cite{adversarial_example}. This vulnerability of DNNs has led to
the proposal of several methods to defend 
adversarial attacks~\cite{distillation_defense, virtual_defense, defense_gan_defense, trades_defense}. Despite these attempts, many of these defenses have been defeated by strong adversarial attacks~\cite{bim_attack, pgd_attack, cw_attack}, or were eventually found to rely on obfuscated gradients~\cite{obfuscated_adversarial}.

Adversarial training~\cite{pgd_attack} is one of the most effective adversarial defense methods which substitutes adversarial examples for the training samples. Given a dataset $D = \{(\bm{\mathnormal{x}}_i, {\mathnormal{y}}_i)\}_{i=1}^n$ with $\bm{\mathnormal{x}}_i \in \mathbb{R}^d$ as an example in the $d$-dimensional input space and $\mathnormal{y}_i$ as its associated label, the goal of adversarial training is to train models by using adversarial empirical risk minimization~\cite{pgd_attack}:
\begin{equation}
    \min_{\theta}\mathop{\mathbb{E}}_{(\bm{\mathnormal{x}},\mathnormal{y})\mathtt{\sim}D}\left[\max_{\mathbf{\delta} \in S} \mathcal{L}(\bm{\mathnormal{x} + \delta},\mathnormal{y};\theta{}) \right].
\end{equation}


Here, $\mathcal{L}(\bm{\mathnormal{x} + \delta},\mathnormal{y};\theta)$ is the loss function on adversarial examples, and $S$ represents the set of perturbations an adversary can apply to deceive the model, which is normally the set of $\ell_p\text-$bounded perturbations.

Many studies of the properties of these adversarial perturbations have been reported. Gilmer~\etal~\cite{sphere_adversarial} noted that the phenomenon of adversarial examples appears because most high dimensional data points in the data distribution are very close to the points that could be adversarial examples. Schmidt~\etal~\cite{more_data_adversarial} proved that robust training requires significantly larger sample complexity than that of standard training, postulating that the difficulty of robust training originates from the large sample complexity. Tsipras~\etal~\cite{odds_with_accuracy} showed that a trade-off may exist between adversarial robustness and standard accuracy. They argued that the features learned during adversarial training differ from those learned during standard training, and attributed the trade-off to this difference.


 Recently, Ilyas \etal~\cite{not_bugs_are_features} demonstrated that the features used to train deep learning models can be divided into adversarially robust features and non-robust features, and the problem of adversarial examples may arise from these non-robust features. Then, if adversarial examples are features, rather than bugs, it is natural to wonder: \emph{Could we take into account the generalization between \textquotedblleft adversarial features\textquotedblright{} in our adversarial training?} If so, \emph{is the large gap between test accuracy and training accuracy under adversarial perturbations during adversarial training caused by the failure of adversarial feature generalization?} 
 
 Motivated by these questions, we present a theoretical model which demonstrates the robust generalization performance changes during adversarial training. Specifically, we define a generalization problem of adversarial training and show that our proposed method can alleviate the generalization problem. In summary, our paper makes the following contributions:
 
\begin{itemize}[leftmargin=*, nolistsep]
    \item We present a theoretical analysis which demonstrates the extent to which the change in the variance of the feature representations affects the robust generalization.
    \item We uncover \emph{Adversarial Feature Overfitting}~(AFO), the phenomenon of the model overfitting to the adversarial features during adversarial training leading to poor robust generalization.
    \item We propose \emph{Adversarial Vertex mixup}~(AVmixup), a soft-labeled data augmentation approach for adversarial training in a collaborative fashion.
    \item We analyze our proposed method with the results of experiments on CIFAR10, CIFAR100, SVHN, and Tiny Imagenet, and show that AVmixup substantially increases the effectiveness of state-of-the-art adversarial training methods.
\end{itemize}
\section{Background} 
\label{Background}

\subsection{Adversarially Robust Generalization} \label{3.1}



Schmidt~\etal~\cite{more_data_adversarial} showed that the sample complexity for robust generalization can be much larger than the sample complexity for standard generalization by constructing a toy example as follows:

\theoremstyle{plain}
\newtheorem{exmp}{Example}
\begin{exmp} 
\label{exmp1}

\textup{(Schmidt~\etal)} Let $\theta^{\star{}} \in \mathbb{R}^d$ be the per-class mean vector and let $\sigma > 0$ be the variance parameter. Then the $(\theta^{\star{}}, \sigma)\text-$Gaussian model is defined by the following distribution over $(\bm{\mathnormal{x}}, \mathnormal{y}) \in \mathbb{R}^d \times \{\pm1\}$
\begin{equation}
    \mathnormal{y}\stackrel{u.a.r.}{\sim} \{-1, +1\},\;\;
    \bm{\mathnormal{x}}\stackrel{i.i.d.}{\sim}\mathcal{N}(\mathnormal{y}\cdot\theta^{\star}, \sigma^{2}\mathnormal{I}).
\end{equation}

\end{exmp}
Here, the difficulty of the binary classification task is controlled by adjusting the variance parameter $\sigma$ which implies the amount of overlap between the two classes. 

To characterize robust generalization, the definitions of standard and robust classification error are defined as follows (Schmidt~\etal):
\newtheorem{definition}{Definition}
\begin{definition} \label{def1}
    Let $Q:\mathbb{R}^d\times\{\pm1\}\xrightarrow{}\mathbb{R}$ be a distribution. Then the standard classification error $\beta{}$ of a classifier $\mathnormal{f}:\mathbb{R}^d\xrightarrow{}\{\pm1\}$ is defined as $\beta=\mathop{\mathbb{P}}_{(\bm{\mathnormal{x}},\mathnormal{y})\mathtt{\sim}Q}[\mathnormal{f}(\bm{\mathnormal{x}})\neq{}\mathnormal{y}]$.
\end{definition}

\begin{definition} \label{def2}
    Let $Q:\mathbb{R}^d\times\{\pm1\}\xrightarrow{}\mathbb{R}$ be a distribution and let $S\in\mathbb{R}^d$ be a perturbation set that the adversary could apply to fool the model. Then the $S\text-$robust classification error $\beta{}$ of a classifier $\mathnormal{f}:\mathbb{R}^d\xrightarrow{}\{\pm1\}$ is defined as $\beta=\mathop{\mathbb{P}}_{(\bm{\mathnormal{x}},\mathnormal{y})\mathtt{\sim}Q}[\exists\bm{\delta} \in S:\mathnormal{f}(\bm{\mathnormal{x}+\delta})\neq{}\mathnormal{y}]$.
\end{definition}
Hence, the $\ell^{\epsilon}_p\text-$robustness is defined as robustness with respect to the perturbation set $S=\{ \bm{\delta} \in \mathbb{R}^d \mid \| \delta \|_{p}\leq\epsilon \}$. In our work, we focus on $\ell_\infty\text-$bounded perturbations, because this is the most common type in the context of adversarial perturbations~\cite{pgd_attack,bim_attack, trades_defense, feature_scattering_defense}.


To calculate the sample complexities for robust and standard generalization, Schmidt~\etal~\cite{more_data_adversarial} used the following linear classifier model:

\begin{definition}
    \label{def3}
    \textup{(Schmidt~\etal)} Let $(\bm{\mathnormal{x}}_1, \mathnormal{y}_1),\dots,(\bm{\mathnormal{x}}_n, \mathnormal{y}_n) \in \mathbb{R}^d\times\{\pm1\}$ be drawn i.i.d. from a $(\theta^{\star{}}, \sigma)\text-$Gaussian model with $\|\theta^\star{}\|_2=\sqrt{d}$. Let the weight vector $\bm{\mathnormal{w}} \in \mathbb{R}^d$ be the unit vector in the direction of $\Bar{\bm{\mathnormal{z}}}=\frac{1}{n}\sum^{n}_{i=1}\mathnormal{y}_i\bm{\mathnormal{x}}_i$.
    Then the linear classifier $\mathnormal{f}_{n,\sigma}$ is defined as
    \begin{equation}
        \mathnormal{f}_{n,\sigma} = \text{sgn}(\bm{\mathnormal{w}}^\top \bm{\mathnormal{x}}).
    \end{equation}
\end{definition} 

It was shown that the linear classifier can achieve satisfactory generalization performance even with a single sample when the variance of the data distribution is small. The upper $\ell_\infty\text-$bound of adversarial perturbations was also derived for a certain $\ell^{\epsilon}_\infty\text-$robust classification error under the same conditions with standard classification.

\subsection{Robust and Non-robust Features}



Recent studies~\cite{odds_with_accuracy, not_bugs_are_features} considered the adversarial robustness in the existence of a distinction between robust features and non-robust features. They noted that adversarial examples can arise from the non-robust features of input data which are useful for standard classification but have an adverse effect on robust classification~\cite{not_bugs_are_features}. They provided evidence to support the hypothesis by showing that non-robust features alone are sufficient for standard classification but not for robust classification. They also demonstrated that standard training on the set of robust features yields a fairly small robust classification error.


Tsipras~\etal~\cite{odds_with_accuracy} indicated that the existence of a provable trade-off between standard accuracy and its robustness. They theoretically showed the possibility that adversarial robustness is incompatible with standard accuracy in a simple setting using a Gaussian model. In addition, they emphasized that adversarial training may reduce the contribution of non-robust features to zero with the following lemma:
\newtheorem{lemma}{Lemma}
\begin{lemma}
\label{lemma}
    \textup{(Tsipras~\etal)} Minimizing the adversarial empirical risk results in a classifier that assigns 0 weight to non-robust features.
\end{lemma}

\subsection{Soft Labeling}
Szegedy~\etal~\cite{inception} proposed label-smoothing as a mechanism to regularize the classifier. They argued that maximizing the log-likelihood of the correct label may result in overfitting, and label-smoothing can alleviate the overfitting problem.

Zhang~\etal~\cite{mixup} introduced a novel data augmentation method named Mixup. Mixup constructs virtual training examples as follows: 
\begin{equation}
    \Tilde{\mathnormal{x}}=\alpha\mathnormal{x}_i+(1-\alpha)\mathnormal{x}_j,\quad
    \Tilde{\mathnormal{y}}=\alpha\mathnormal{y}_i+(1-\alpha)\mathnormal{y}_j.
\end{equation}
$(\mathnormal{x}_i,\mathnormal{y}_i)$ and $(\mathnormal{x}_j,\mathnormal{y}_j)$ are two examples drawn at random from the training data, and $\alpha \in [0,1]$.
They showed that Mixup improves generalization on various tasks.

\section{Methods}

\subsection{Theoretical Motivation} \label{theorems}
In this section, we theoretically analyze the statistical aspects of robust generalization. First, a simple Gaussian data model is used to demonstrate the need to minimize feature representation variance for robust generalization. It is then shown that the optimal model parameter in terms of robust generalization differs from the model parameter which minimizes the adversarial empirical risk using data which consist of robust and non-robust features. Ultimately, we provide evidence that most deep neural networks are not free from AFO by showing that even in our simple Gaussian data model, the robust generalization performance is degraded as the model is overly trained on adversarial examples.

Based on Example~\ref{exmp1} and the linear classifier defined in Definition~\ref{def3}, we prove the following theorem:

\newtheorem{theorem}{Theorem}
\begin{theorem} \label{theorem1}
    For the variance parameters $\sigma_r$ and $\sigma_s$~(subscript r for robust and s for standard), let $\sigma_r=\nu\sigma_s$ where $\nu \in [0,1]$. Then, the upper bound on the standard classification error of $\mathnormal{f}_{n,\sigma_s}$ and the upper bound on the $\ell^{\epsilon}_\infty\text-$robust classification error of $\mathnormal{f}_{n,\sigma_r}$ be equal with probability at least $\left( 1-2\exp(-\frac{d}{8(\sigma_s^2+1)}) \right)\cdot\left( 1-2\exp(-\frac{d}{8(\sigma_r^2+1)}) \right)$ if
    \begin{equation}
        \epsilon \leq \frac{(2\sqrt{n}-1)(1-\nu)}{2\sqrt{n}+4\sigma_s}.
    \end{equation}
\end{theorem}


\noindent
(All the proofs of the theorems and corollaries in our work can be found in the supplementary material.) We can see that the theorem is consistent with our intuition. For example, when $\nu=1$, \ie, when both variances are equal, the probability that the robust generalization ability for $\epsilon>0$ is the same as the standard generalization ability effectively becomes zero. Thus, to ensure that our model shows robust generalization at the same level as standard generalization, a smaller variance of feature representations is required than that of standard learning.
\newtheorem{corollary}{Corollary}
\begin{corollary}
    For the variance parameters $\sigma_r$ and $\sigma_s$, let $\sigma_r=\nu\sigma_s$ where $\nu \in [0,1]$. Let the upper bound on the standard classification error of $\mathnormal{f}_{n,\sigma_s}$ and the upper bound on the $\ell^{\epsilon}_\infty\text-$robust classification error of $\mathnormal{f}_{n,\sigma_r}$ be equal. Then, as $\sigma_r$ decreases, the upper bound of $\epsilon$ increases in proportion to $\pi_{n,\sigma_s}$, which is given by
    \begin{equation}
        \pi_{n,\sigma_s}=\frac{2\sqrt{n}-1}{\sigma_s(2\sqrt{n}+4\sigma_s)}.
    \end{equation}
\end{corollary}

Hence, the smaller the variance of feature representations, the more effective the robust generalization performance of the model. 

Next, we show the change in the variance of feature representations as we train the model to minimize the adversarial empirical risk. Specifically, we utilize the concept of robust and non-robust features, and show the way in which adversarial training results in AFO in a model similar to that used before~\cite{odds_with_accuracy}.

\begin{exmp}
\label{exmp2}

Let $0<\sigma_A\ll\sigma_B$. Then, the distribution $\Psi_{true}$ is defined by the following distribution over $(\bm{\mathnormal{x}}, \mathnormal{y}) \in \mathbb{R}^{d+1} \times \{\pm1\}:$
\begin{equation}
    \begin{gathered}
    \mathnormal{y}\stackrel{u.a.r}{\sim} \{-1, +1\} \quad \text{and} \\
    \mathnormal{x}_1\stackrel{}{\sim}\mathcal{N}(\mathnormal{y},\sigma_{A}^2),
    \quad\mathnormal{x}_2,\dots,\mathnormal{x}_{d+1}\stackrel{i.i.d.}{\sim}\mathcal{N}(\eta\mathnormal{y}, \sigma_{B}^2).
    \end{gathered}
\end{equation}

\end{exmp}
Here, $\mathnormal{x}_1$ is a robust feature that is strongly correlated with the label, and the other features $\mathnormal{x}_2,\dots,\mathnormal{x}_{d+1}$ are non-robust features that are weakly correlated with the label. Here, $\eta<1$ is a non-negative constant, which is small but sufficiently large such that a simple classifier attains a small standard classification error.


The difficulty associated with robust learning is that a significantly large sample complexity is required~\cite{more_data_adversarial}. Given this postulation, we extend Example~\ref{exmp2} to Example~\ref{exmp3} with the following assumption:

\newtheorem{assumption}{Assumption}
\begin{assumption} \label{assumption}

    Assume the number of non-robust features in our data is $N$. Then, because of the lack of data samples in robust learning, $M$ features out of $N$ non-robust features form a sample distribution which is far from the true distribution.
    
\end{assumption}

In Assumption~\ref{assumption}, we refer to $M$ non-robust features as \textquotedblleft insufficient\textquotedblright{} non-robust features.
Contrarily, the other non-robust features are referred to as \textquotedblleft sufficient \textquotedblright{} non-robust features.

\begin{exmp}
\label{exmp3}
Let $0<c<d$. Then
the sample distribution $\Psi_{sample,c}$ which is formed by the sampled input-label pairs $(\bm{\mathnormal{x}},\mathnormal{y})\stackrel{i.i.d.}{\sim}\Psi_{true}$ is defined as follows:
\begin{equation}
    \begin{gathered}
    \mathnormal{y}\stackrel{u.a.r.}{\sim} \{-1, +1\}, 
    \quad\mathnormal{x}_1\stackrel{}{\sim}\mathcal{N}(\mathnormal{y},\sigma_{A}^2), \\
    \mathnormal{x}_2,\dots,\mathnormal{x}_{c+1}\stackrel{i.i.d.}{\sim}\mathcal{N}(\mathnormal{y}, \sigma_{A}^2), \\
    \mathnormal{x}_{c+2},\dots,\mathnormal{x}_{d+1}\stackrel{i.i.d.}{\sim}\mathcal{N}(\eta\mathnormal{y}, \sigma_{B}^2).
    \end{gathered}
\end{equation}

\end{exmp}


In Example~\ref{exmp3}, our data has a true distribution as in Example~\ref{exmp2}. However, the Gaussian distribution is changed for the insufficient non-robust features $\mathnormal{x}_2,\dots,\mathnormal{x}_{c+1}$ in our sampled data according to Assumption~\ref{assumption}. For simplicity, in this example, we suppose that the insufficient non-robust features form the same sample distribution as that of the robust features.


We show the variance of feature representations during adversarial training on $\Psi_{sample,c}$ by using the following linear classifier:

\begin{definition}

    \label{def4}
    Let $Z$ be a function set. Let $\bm{\mathnormal{w}}$ be the weight vector of the classifier. Let $\zeta_f$ be the objective function of the linear classifier $\mathnormal{f}_{\bm{\mathnormal{w}}}$.
    Then our linear classifier $\mathnormal{f}_{Z}$ is defined as
    \begin{equation}
        \mathnormal{f}_{Z} \in \{\mathnormal{f}_{\bm{\mathnormal{w}}} \mid \zeta_f \in Z, 
        \bm{\mathnormal{w}} \in \mathbb{R}_{+}^{d+1}, 
        \|\bm{\mathnormal{w}}\|_1=1\}.
    \end{equation}

\end{definition} 

In our model, it is reasonable to investigate the variance of $\bm{\mathnormal{w}}^\top\bm{\mathnormal{x}}$ to show the extent to which adversarial training affects robust generalization. Based on Example~\ref{exmp3} and the linear classifier defined in Definition~\ref{def4}, we can prove the following theorem:

\begin{theorem}
\label{theorem2}
    Let $\bm{\mathnormal{w}}_B \in \mathbb{R}^{d-c}$ be the weight vector for the sufficient non-robust features of $\Psi_{sample,c}$. Let $Z_{sc}$ be a set of strictly convex functions. Then, when the classifier $\mathnormal{f}_{Z_{sc}}$ is trained on $\Psi_{sample,c}$, the $\bm{\mathnormal{w}}^{\star}_B$ which minimizes the variance of $\bm{\mathnormal{w}}^\top\bm{\mathnormal{x}}$ with respect to $\Psi_{sample,c}$ is
    \begin{equation}
        \bm{\mathnormal{w}}^{\star}_B = \Vec{0}.
    \end{equation}
\end{theorem}



This result is consistent with that of~\cite{odds_with_accuracy}, which presumed that the number of samples for all the non-robust features is sufficiently large.
However, we have a limited number of samples for the non-robust features in Example \ref{exmp3} and this may cause the result to differ from that of Theorem~\ref{theorem2}.
Therefore, we need to find $\bm{\mathnormal{w}}^\star_B$ with respect to the true distribution $\Psi_{true}$ for the purpose of showing robust generalization ability for our model.

\begin{theorem}
\label{theorem3}
    Let $\bm{\mathnormal{w}}_B \in \mathbb{R}^{d-c}$ be the weight vector for sufficient non-robust features of $\Psi_{sample,c}$. Let $Z_{sc}$ be a set of strictly convex functions. Then, when the classifier $\mathnormal{f}_{Z_{sc}}$ is trained on $\Psi_{sample,c}$, the $\bm{\mathnormal{w}}^{\star}_B$ that minimizes the variance of $\bm{\mathnormal{w}}^\top\bm{\mathnormal{x}}$ with respect to $\Psi_{true}$ is
    \begin{equation}
        \bm{\mathnormal{w}}^{\star}_B = \frac{c}{cd+2c+1} \cdot \Vec{1}.
    \end{equation}
\end{theorem}


For simplicity, we assume that the classifier assigns the same weight value to features with the same distribution in Theorem~\ref{theorem3}, and the limited feasible set does not change the optimal weight of the classifier.
As a result, we can predict the robust generalization performance of the classifier by observing $\bm{\mathnormal{w}}_B$ in the robust learning procedure. 
Note that Lemma~\ref{lemma} also applies with our classifier. Therefore,
if our sampled data have insufficient non-robust features, $\bm{\mathnormal{w}}_B$ approaches $\vec{0}$ during adversarial training, even though the optimal $\bm{\mathnormal{w}}^\star_B$ is not $\vec{0}$ in terms of robust generalization. We refer to this phenomenon as \emph{Adversarial Feature Overfitting~(AFO)}.

AFO is caused by the relation between the weight values for the features in our data. In this regard, most deep neural networks involve intertwined features, suggesting that most deep neural networks are also adversely affected by the problem we point out in the example.

\subsection{Adversarial Vertex Mixup}

\begin{figure}[t]
\begin{center}
\includegraphics[width=0.8\linewidth]{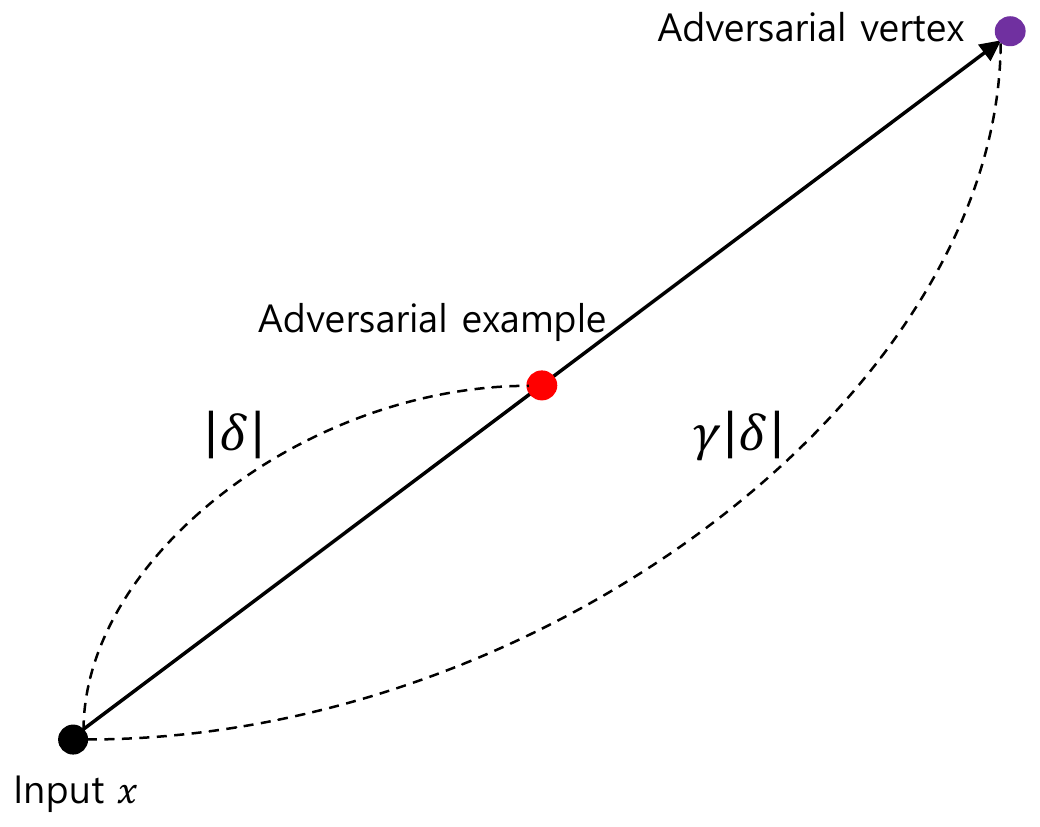}
\end{center}
   \caption{\textbf{Adversarial vertex}. 
   The adversarial vertex is located in the same direction as the adversarial example but $\gamma$ times farther away.}
\label{fig1}
\end{figure}


AFO arises when the model is overly optimized only for sufficient non-robust features, when the training data have many types of insufficient non-robust features. From this point of view, we can think of several methods to address AFO. First, the diversity of the algorithm that constructs adversarial examples during training could be increased. This may be a fundamental solution to overcome the poor robust generalization caused by the large sample complexity. Second, when the large sample complexity of robust learning cannot be satisfied, label-smoothing can directly regularize the overfitting problem as in~\cite{inception}. Essentially, soft labeling can be employed to prevent the weights for the sufficient non-robust features from becoming zero. In this paper, we present a method to improve the robust generalization using soft labeling.


Several algorithms that use soft-labeled data to improve the generalization performance have been proposed~\cite{inception, mixup, soft_label_reg}. Among them, Mixup~\cite{mixup} trains a model by utilizing linear interpolation between training data. This method can be seen as a variant of the label-smoothing method, because it linearly interpolates both input vectors and their labels. Inspired by Mixup, we propose Adversarial Vertex mixup~(AVmixup), which is a soft-labeled data augmentation method designed to improve robust generalization.


AVmixup, similar to Mixup, also extends the training distribution by using linear interpolation. 
Unlike Mixup, however, for each raw input vector, AVmixup defines a virtual vector in the adversarial direction and extends the training distribution via linear interpolation of the virtual vector and the raw input vector.
We refer to the virtual vector as an \emph{adversarial vertex}~(see Figure~\ref{fig1}). Formally, the adversarial vertex is defined as follows:

\begin{definition}

    Let $\delta_{\bm{\mathnormal{x}}} \in \mathbb{R}^d$ be the adversarial perturbation for the raw input vector $\bm{\mathnormal{x}} \in \mathbb{R}^d$.
    Then, for a scaling factor $\gamma \geq 1$, adversarial vertex $\bm{\mathnormal{x}}_{av}$ is defined as
    \begin{equation}
        \bm{\mathnormal{x}}_{av}=\bm{\mathnormal{x}}+\gamma \cdot \delta_{\bm{\mathnormal{x}}}.
    \end{equation}
    
\end{definition}


Figure~\ref{fig1} shows how the adversarial vertex is found. After we obtain the adversarial vertex, AVmixup constructs virtual training examples as follows:

\begin{definition}

    Let $(\bm{\mathnormal{x}}, \bm{\mathnormal{y}})$ be the raw input-label pair.
    Let $\phi$ be a label-smoothing function. 
    Then, for the real value $\alpha$ sampled from a uniform distribution $\mathcal{U}(0,1)$ and the label-smoothing parameters $\lambda_1 \in \mathbb{R}$ and $\lambda_2 \in \mathbb{R}$, the virtual input vector $\hat{\bm{\mathnormal{x}}} \in \mathbb{R}^d$ and its associated label $\hat{\bm{\mathnormal{y}}} \in \mathbb{R}^k$ are constructed by
    \begin{equation}
        \begin{gathered}
        \hat{\bm{\mathnormal{x}}} = \alpha \bm{\mathnormal{x}} + (1-\alpha)\bm{\mathnormal{x}}_{av}, \\
        \hat{\bm{\mathnormal{y}}} = \alpha \phi(\bm{\mathnormal{y}}, \lambda_1) + (1-\alpha)\phi(\bm{\mathnormal{y}}, \lambda_2).
        \end{gathered}
    \end{equation}

\end{definition}

For the label-smoothing function $\phi$, we use an existing label-smoothing method~\cite{inception}.
Specifically, in the case of $k$ classes, the algorithm assigns $\lambda \in (0,1)$ to the true class and equally distributes $\frac{1-\lambda}{k-1}$ to the other classes. 

In summary, the overall procedure of adversarial training with AVmixup is described in Algorithm~\ref{algo}.

\begin{algorithm}
\caption{Adversarial Training with AVmixup}
\begin{algorithmic}[1]
\label{algo}
\REQUIRE Dataset $D$, batch size $n$, training epochs $T$, learning rate $\tau$, scaling factor $\gamma$, label-smoothing factors $\lambda_1, \lambda_2$
\REQUIRE Label-smoothing function $\phi$
\REQUIRE Adversarial perturbation function $\mathcal{G}$\FOR{$t=1$ \textbf{to} $T$}
\FOR{mini-batch $\{\bm{\mathnormal{x}}_i,\bm{\mathnormal{y}}_i\}^n_{i=1} \sim D$}
\STATE $\bm{\delta}_i \leftarrow \mathcal{G}(\bm{\mathnormal{x}}_i,\bm{\mathnormal{y}}_i;\bm{\theta})$
\STATE \emph{AVmixup}:
\STATE $\Bar{\bm{\mathnormal{x}}}_i \leftarrow \bm{\mathnormal{x}}_i + \gamma \cdot \bm{\delta}_i$,\quad$\alpha_i \sim \mathcal{U}(0,1)$
\STATE $\hat{\bm{\mathnormal{x}}}_i \leftarrow \alpha_i \bm{\mathnormal{x}}_i + (1-\alpha_i)\bar{\bm{\mathnormal{x}}}_i$
\STATE $\hat{\bm{\mathnormal{y}}}_i \leftarrow \alpha_i \phi(\bm{\mathnormal{y}}_i,\lambda_1) + (1-\alpha_i)\phi({\bm{\mathnormal{y}}}_i,\lambda_2)$
\STATE \emph{model update}:
\STATE $\bm{\theta} \leftarrow \bm{\theta} - \tau\cdot\frac{1}{n}\sum^n_{i=1}\nabla_{\theta}\mathcal{L}(\hat{\bm{\mathnormal{x}}}_i,\hat{\bm{\mathnormal{y}}}_i;\bm{\theta})$
\ENDFOR
\ENDFOR
\STATE \textbf{Output:} robust model parameter $\bm{\theta}$
\end{algorithmic}
\end{algorithm}

\section{Related Work}
\subsection{Adversarial Attack Methods}

Adversarial attacks confuse the trained deep neural networks with adversarial examples. The Fast Gradient Sign Method~(FGSM)~\cite{fgsm_attack} is an efficient one-step attack method. Projected Gradient Descent~(PGD)~\cite{pgd_attack} constructs adversarial examples by applying a multi-step variant of FGSM. The Carlini \& Wagner~(CW) attack~\cite{cw_attack} uses a specific objective function to create adversarial examples under various conditions. Apart from these attacks, many adversarial attacks exist in white-box settings~\cite{jsma_attack,deepfool_attack,bim_attack,universal_attack}. In black-box settings, adversarial attacks are conducted using substitute models, according to which adversarial examples are generated from the substitute models~\cite{substitute_attack}. Additionally, black-box attacks which only rely on the prediction score or the decision of the model have been proposed~\cite{zoo_attack, boundary_attack, singlepixel_attack, nes_attack, simba_attack}. 

\subsection{Adversarial Defense Methods}
Various adversarial defense methods have been employed to make DNNs robust to adversarial attacks. Adversarial training~\cite{fgsm_attack} uses adversarial examples as training data to train the robust network. 
Many approaches~\cite{alp_defense,virtual_defense,double_backprop_defense,trades_defense,feature_scattering_defense} improve the model robustness through regularizers or variants of adversarial training. Various techniques~\cite{detecting_defense, magnet_defense, feature_squeezing_defense, defense_gan_defense, feature_denoising_defense} can defend adversarial attacks by denoising adversarial perturbations from input data or detect adversarial examples from among the input data. 
We cover further related works in the supplementary material.
\section{Experimental Results And Discussion}


In this section, we show that label-smoothing~\cite{inception} and AVmixup improve the robust generalization with extensive experiments across many benchmark datasets including CIFAR10~\cite{cifar_dataset}, CIFAR100~\cite{cifar_dataset}, SVHN~\cite{svhn_dataset} and Tiny Imagenet~\cite{imagenet_dataset}. Especially, we note that the combination of AVmixup with the state-of-the-art adversarial defense method~\cite{feature_scattering_defense}, would enable us to \emph{significantly outperform existing defense methods}. A description of the datasets used in the experiments is summarized in the supplementary material.

\subsection{Implementation Details}

We use
WRN-34-10~\cite{wide_resnet} for the experiments on CIFAR, WRN-16-8~\cite{wide_resnet} for the experiments on SVHN, and PreActResNet18~\cite{preact_resnet} for the experiments on Tiny Imagenet. We run 80k training steps on CIFAR and SVHN and 50k training steps on Tiny Imagenet. The initial learning rate for CIFAR and Tiny Imagenet is set to 0.1 and 0.01 for SVHN. The learning rate decay is applied at 50\% and 75\% of total training steps with decay factor 0.1, and weight decay factor is set to $2\mathrm{e}{-4}$. We use the same adversarial perturbation budget $\epsilon=8$ as in~\cite{pgd_attack}. To evaluate adversarial defense methods, we apply several adversarial attacks including FGSM~\cite{fgsm_attack}, PGD~\cite{pgd_attack}, CW~\cite{cw_attack}~(PGD approach with CW loss) and transfer-based black-box attack~\cite{substitute_attack}.
We mainly compare the following settings in our experiments:
\begin{enumerate}[leftmargin=*, nolistsep]
    \item Standard: The model which is trained with the original dataset. 
    \item PGD: The model trained using adversarial examples from PGD~\cite{pgd_attack} with $\textup{step size}=2$, $\textup{iterative steps}=10$.
    \item LS$\lambda$: With the PGD-based approach~\cite{pgd_attack}, we apply the label-smoothing method~\cite{inception} for the model with label-smoothing factor $\lambda$.
    \item AVmixup: We apply our proposed method for the model with the PGD-based approach~\cite{pgd_attack}.
\end{enumerate}
Note that PGD and CW attacks with $T$ iterative steps are denoted as PGD$T$ and CW$T$, respectively, and the original test set is denoted as Clean.

\subsection{CIFAR10}
\label{cifar10}

\begin{table}[]
\centering
\caption{Comparison of the accuracy of our proposed approach AVmixup with that of PGD~\cite{pgd_attack} and LS$\lambda$~($\lambda \in \{0.8,0.9\}$)~\cite{inception} against white-box attacks on CIFAR10.}
\begin{tabular}{cccccc}
\toprule
\multicolumn{1}{c|}{Model} & Clean                & FGSM                 & PGD10                & PGD20                & CW20                 \\ \hline
\multicolumn{1}{c|}{Standard}    & 95.48                & 7.25                     & 0.0                & 0.0                & 0.0                \\
\multicolumn{1}{c|}{PGD}    & 86.88                & 62.68                      & 47.69                & 46.34                & 47.35                \\
\multicolumn{1}{c|}{LS0.8}  & 87.28                & 66.09                      & 53.49                & 50.87                & 50.60                 \\
\multicolumn{1}{c|}{LS0.9}  & 87.64                & 65.96                      & 52.82                & 50.29                & 50.30                 \\
\multicolumn{1}{c|}{AVmixup} & \textbf{93.24}       & \textbf{78.25}                      & \textbf{62.67}       & \textbf{58.23}       & \textbf{53.63}       \\ \bottomrule
\end{tabular}
\label{tab:table-1}
\end{table}

\begin{figure}[t]
\begin{center}
\includegraphics[width=\linewidth]{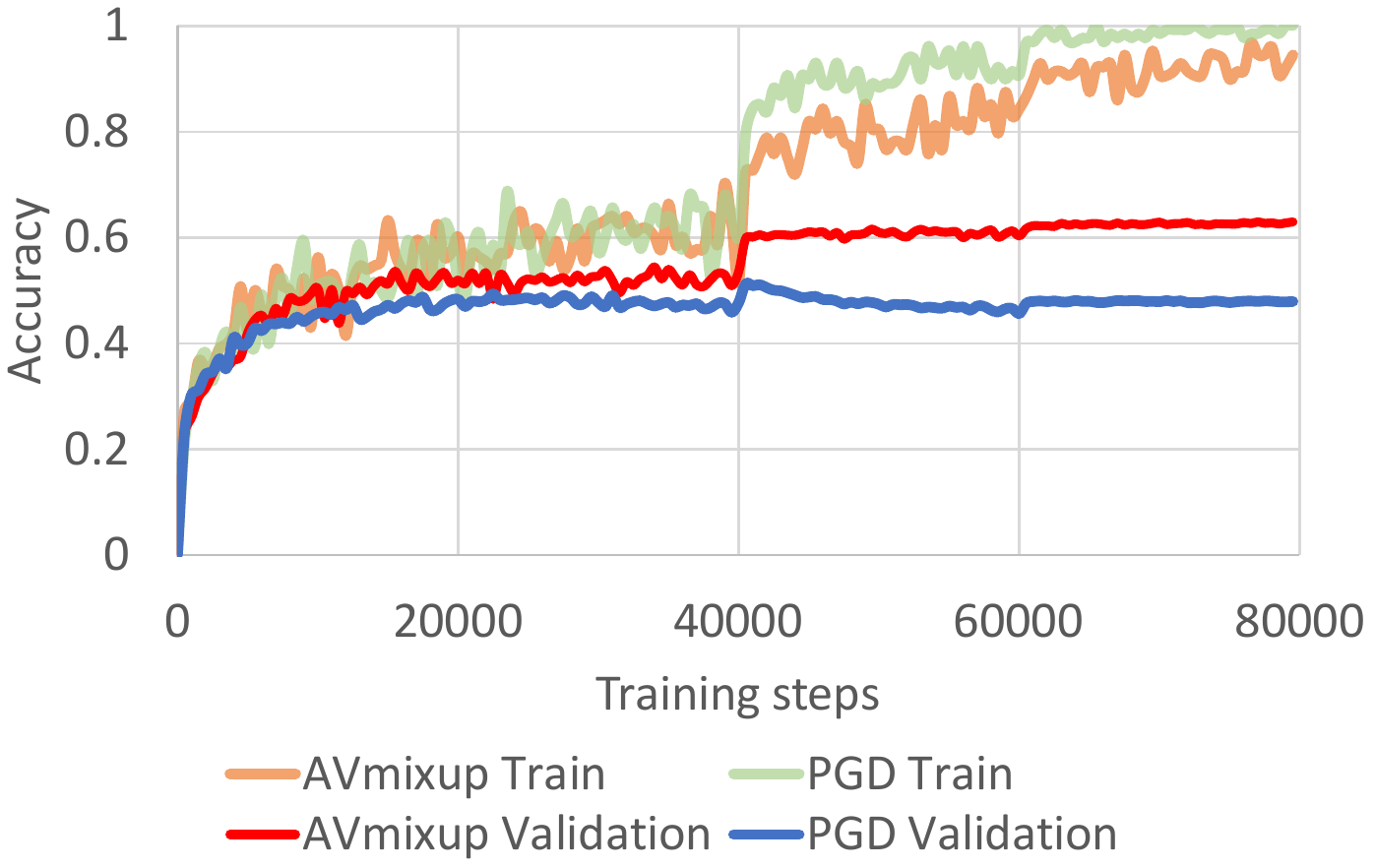}
\end{center}
   \caption{\textbf{CIFAR10 accuracy curves}. 
   The robustness of the PGD model~(blue line) overfits after 40k steps. The AVmixup model, on the other hand, shows a steady increase in robustness~(red line).}
\label{fig2}
\end{figure}

Because the CIFAR10 dataset is the most commonly used dataset for adversarial robustness studies~\cite{pgd_attack, feature_squeezing_defense, trades_defense, feature_scattering_defense}, we analyze our method in both white-box and black-box settings, and compare our method to a state-of-the-art defense method, TRADES~\cite{trades_defense}, on CIFAR10. We set the scaling factor $\gamma=2.0$ and label-smoothing factors $\lambda_1=1.0$ and $\lambda_2=0.1$ in the following experiments.

\paragraph{Empirical evidence for AFO}
We provide Figure~\ref{fig2} in support of our theoretical analysis and the effectiveness of AVmixup.
In Figure~\ref{fig2}, the validation accuracy curve against PGD10 of the PGD model shows that the model starts to overfit from about 40k steps, while the AVmixup model continues to improve.

\paragraph{White-box setting}\label{white-box}

We conduct white-box attacks on the models trained with baseline methods and our proposed method AVmixup. We set the $\textup{step size}=2.0$ for PGD and CW attacks.
We first evaluate the models on Clean to compare the trade-off between accuracy and robustness of the models.
Then, we evaluate the models against FGSM, PGD10, PGD20, and CW20. The results are summarized in Table~\ref{tab:table-1}. 
    
The results in Table~\ref{tab:table-1} indicate that models trained with soft labels are more accurate in all attacks including clean data than the model trained with one-hot labels, which is consistent with our theoretical analysis. In particular, the accuracy on PGD20 of the AVmixup model is 11.89\%p higher than that of the PGD model, with a decrease of only 2.24\%p in accuracy on Clean compared to the Standard model. 

\paragraph{Black-box setting}
Athalye~\etal~\cite{obfuscated_adversarial} indicated that 
obfuscated gradients, a phenomenon that leads to non-true adversarial defenses, can be identified in several ways. One such way is black-box attack evaluation.
    
In black-box settings, we apply transfer-based black-box attacks to the models~\cite{substitute_attack}. After constructing adversarial examples from each of the trained models, we apply these adversarial examples to the other models and evaluate the performances. The results are summarized in Table~\ref{tab:table-2}, and more results can be found in the supplementary material. The columns represent the attack models of the transfer-based black-box attacks, and the rows represent the defense models which are evaluated. The results in Table~\ref{tab:table-2} indicate that the AVmixup model is the most robust against black-box attacks from all of the attack models with significant margins.
We also observe that the model trained with AVmixup shows higher accuracy against black-box attacks than against white-box attacks. Thus, we confirm that our proposed method improves the adversarial defense performance as a result of an increase in the robustness of the model rather than with obfuscated gradients~\cite{obfuscated_adversarial}.

\paragraph{Comparison}
 We compare our method with a recently proposed defense method, TRADES~\cite{trades_defense}, which uses a regularization-based adversarial training approach.
TRADES requires approximately twice as much GPU memory as conventional adversarial training to calculate the additional regularization term which processes both natural examples and adversarial examples simultaneously. In contrast, AVmixup hardly incurs additional cost and can be implemented with only a few lines of code. In this experiment, we implement AVmixup based on the official PyTorch code of TRADES~\cite{trades_defense} and train the model with the same configurations as~\cite{pgd_attack}. The results are listed in Table~\ref{tab:table-3}, which shows that our proposed method has superior robustness with a smaller trade-off than TRADES.

\paragraph{Discussion} \label{discussion1}
    In Tabel~\ref{tab:table-1}, in contrast with FGSM, PGD10 and PGD20, the AVmixup model does not show significant improvement against the CW20 attack, and this trend becomes more severe for challenging datasets that have a larger number of classes and smaller number of training examples per class such as CIFAR100. We can infer 
    that this property appears as AVmixup uses linear interpolations. In other words, algorithms that utilize virtual data constructed using linear interpolation between data points tightly generalize only the features observed in the training steps. 
    We confirm this explanation by a simple experiment, the details and further discussion of which can be found in the supplementary material.
    It implies that while AVmixup shows a high level of robustness against adversarial attacks used in adversarial training, it may not be able to withstand other types of attacks. Therefore, the diversity of adversarial examples generated during the adversarial training procedure is even more important for AVmixup. We thus report the results of AVmixup combined with the PGD-based approach by focusing on PGD-based attacks. The results using an algorithm that constructs diverse adversarial examples are discussed in Section~\ref{diversity}.

\begin{table}[]
\centering
\caption{Accuracy comparisons against transfer-based black-box attacks~(PGD20).}
\begin{tabular}{c|cccc}
\toprule
\multirow{2}{*}{\begin{tabular}[c]{@{}c@{}}Defense\\ model\end{tabular}} & \multicolumn{4}{c}{Attack model}                                  \\ \cline{2-5} 
                                                                         & Standard       & PGD            & LS0.8          & LS0.9          \\ \hline
PGD                                                                      & 85.6           & -              & 65.70          & 64.91          \\
LS0.8                                                                    & 86.03          & 63.60          & -              & 64.83          \\
LS0.9                                                                    & 86.40          & 63.74          & 65.78          & -              \\
AVmixup                                                                  & \textbf{89.53} & \textbf{68.51} & \textbf{71.48} & \textbf{70.50} \\ \bottomrule
\end{tabular}
\label{tab:table-2}
\end{table}

\begin{table}[]
\centering
\caption{Accuracy comparisons with TRADES~\cite{trades_defense}.}
\begin{tabular}{ccc}
\hline
\multicolumn{1}{c|}{Models} & Clean         & PGD20                \\ \hline
\multicolumn{1}{c|}{PGD~\cite{trades_defense}}    & 87.3                 & 47.04                \\
\multicolumn{1}{c|}{TRADES~($1 / \lambda = 1$)~\cite{trades_defense}} & 88.64                & 49.14                \\
\multicolumn{1}{c|}{TRADES~($1 / \lambda = 6$)~\cite{trades_defense}} & 84.92                & 56.61                \\
\multicolumn{1}{c|}{AVmixup} & \textbf{90.36}        & \textbf{58.27}       \\ \bottomrule

\end{tabular}
\label{tab:table-3}
\end{table}

\begin{table}[]
\centering
\caption{Comparisons of AVmixup on SVHN~\cite{svhn_dataset}, CIFAR100~\cite{cifar_dataset}, and Tiny ImageNet~\cite{imagenet_dataset}.}
\begin{tabular}{c|>{}c >{}c >{}c >{}c}
    \toprule
Dataset & Model & Clean & FGSM & PGD20\\
         \midrule\midrule[.1em]
         \multirow{4}*{\centering CIFAR100}
        & PGD & 61.29 & 46.01 &25.17 \\
        & LS0.8 & 62.1 & 52.33 &28.81 \\
        & LS0.9 & 61.77 & 53.17 &27.13 \\
        & AVmixup & \textbf{74.81} & \textbf{62.76} &\textbf{38.49} \\
        \midrule[.1em]
     \multirow{4}*{SVHN}
        & PGD  & 92.4 & 75.31 & 58.22  \\
        & LS0.8 & 92.15& 75.84 & 59.75 \\
        & LS0.9 & 92.34& 76.14 & 59.28 \\
        & AVmixup & \textbf{95.59} & \textbf{81.83} &\textbf{61.90} \\
    \midrule[.1em]
       \multirow{4}*{Tiny ImageNet}
        & PGD & 41.67 & 20.30 & 13.14 \\
        & LS0.8 & 42.89 & 22.75 & 15.43 \\
        & LS0.9 & 41.71 & 20.96 & 14.03  \\
        & AVmixup & \textbf{54.27} & \textbf{35.46} & \textbf{20.31} \\

    \bottomrule
\end{tabular}
\label{tab:table-4}
\end{table}
    
\begin{table*}[]
\centering
\caption{Comparisons of AVmixup with feature scattering-based approach~\cite{feature_scattering_defense}. For PGD, we refer to the accuracy of~\cite{feature_scattering_defense}. For Feature Scatter, we reproduce and evaluate the model at the end of the training.}
\begin{tabular*}{\textwidth}{c @{\extracolsep{\fill}} ccccccc}
\toprule
Dataset                   & Model                    & Clean          & FGSM           & PGD20          & PGD100         & CW20           & CW100         \\ \midrule[.1em]
\multirow{3}{*}{CIFAR10}  
                          & \multicolumn{1}{|c}{PGD~\cite{feature_scattering_defense}}                      & 85.7           & 54.9           & 44.9           & 44.8           & 45.7           & 45.4          \\
                          & \multicolumn{1}{|c}{Feature Scatter}          & 90.22           & 78.19           & 69.74           & 67.35           & 60.77           & 58.29          \\
                          & \multicolumn{1}{|c}{Feature Scatter + AVmixup} & \textbf{92.37} & \textbf{83.49} & \textbf{82.31} & \textbf{81.88} & \textbf{71.88} & \textbf{69.50} \\ \midrule[.05em]
                          \multirow{3}{*}{CIFAR100} 
                          & \multicolumn{1}{|c}{PGD~\cite{feature_scattering_defense}}                      & 59.9           & 28.5           & 22.6           & 22.3           & 23.2           & 23.0          \\
                          & \multicolumn{1}{|c}{Feature Scatter}          & 74.9           & 72.99           & 45.29           & 42.77           & 27.35           & 24.89          \\
                          & \multicolumn{1}{|c}{Feature Scatter + AVmixup} & \textbf{78.62}   & \textbf{78.92}   & \textbf{47.28}   & \textbf{46.29}   & \textbf{33.20}   & \textbf{31.22}  \\
 \midrule[.05em]
\multirow{3}{*}{SVHN}     
                          & \multicolumn{1}{|c}{PGD~\cite{feature_scattering_defense}}                      & 93.9           & 68.4           & 47.9           & 46.0           & 48.7           & 47.3          \\
                          & \multicolumn{1}{|c}{Feature Scatter}          & \textbf{96.42}           & \textbf{95.92}           & 58.67           & 46.98           & 51.23           & 38.89          \\
                          & \multicolumn{1}{|c}{Feature Scatter + AVmixup} & 96.07 & 95.26 & \textbf{73.65} & \textbf{70.24}   & \textbf{67.06} & \textbf{62.01} \\ \midrule[.1em]
 
\end{tabular*}
\label{tab:table-5}
\end{table*}

\begin{table}[]
\centering
\caption{Sensitivity of the combination of AVmixup and Feature Scatter to label-smoothing factors~($\gamma=1$) on CIFAR10.}
\begin{tabular}{ccccc}
\toprule
\multicolumn{1}{c|}{$\lambda_1$ / $\lambda_2$} & Clean          & FGSM           & PGD20          & CW20           \\ \hline
\multicolumn{1}{c|}{0.1 / 0.5}    & 91.94          & 80.09          & 74.43          & 62.87          \\
\multicolumn{1}{c|}{0.5 / 0.3}    & 92.82          & 77.81          & 57.67          & 55.18          \\
\multicolumn{1}{c|}{0.5 / 0.5}    & 92.25          & 80.54          & 71.55          & 63.01          \\
\multicolumn{1}{c|}{\textbf{0.5 / 0.7}}    & 92.37          & \textbf{83.49} & \textbf{82.31} & \textbf{71.88} \\
\multicolumn{1}{c|}{0.5 / 1.0}    & 92.11          & 81.75          & 78.70          & 62.96          \\
\multicolumn{1}{c|}{1.0 / 0.5}    & \textbf{93.07} & 79.55          & 53.42          & 56.72          \\ \bottomrule
\end{tabular}
\label{table6}
\end{table}

\subsection{Other Datasets}
We also verify the effectiveness of our method on CIFAR100, SVHN and Tiny Imagenet.
We specify the same hyperparameters for AVmixup as in the CIFAR10 experiments.
The results from these experiments are provided in Table~\ref{tab:table-4}.

\paragraph{CIFAR100}
    Tabel~\ref{tab:table-4} shows that the accuracy of AVmixup increases by 13.52\%p and 13.32\%p for Clean and PGD20, respectively, compared to the PGD model. The results of additional experiments on CIFAR100 can be found in the supplementary material.

\paragraph{SVHN}
    
    The SVHN image classification task is much easier than the image classification tasks with more complicated input images such as CIFAR and Tiny Imagenet. As shown previously~\cite{more_data_adversarial}, generalization problems with poor robustness are less common for simple image datasets such as MNIST than for complex image datasets such as CIFAR. Thus, it is possible to predict that our proposed method, starting from the robust generalization problem, would be less effective on the SVHN dataset than on other datasets, and it is indeed observed from Table~\ref{tab:table-4}. The accuracy of AVmixup improves by 3.19\%p and 3.68\%p compared to the PGD model for Clean and PGD20, respectively, which are small improvements compared to those observed on the other datasets that are tested.

\paragraph{Tiny Imagenet}
    
    Tabel~\ref{tab:table-4} shows an improvement in accuracy of 12.6\%p and 7.17\%p compared to the PGD model for Clean and PGD20, respectively.

\subsection{When AVmixup Meets Diversity}
\label{diversity}

As discussed in~\ref{cifar10}, the diversity of adversarial examples during adversarial training is important to enable AVmixup to be effective against various adversarial attacks. In this sense, we utilize a recent method~\cite{feature_scattering_defense}~(Feature Scatter) which promotes data diversity by taking the inter-sample relationships into consideration during adversarial training. We combine Feature Scatter with our method AVmixup, and evaluate the performance of the model on CIFAR10, CIFAR100 and SVHN.



We implement AVmixup on the PyTorch code of Feature Scatter released in~\cite{feature_scattering_defense}, hence we use the same model architecture and configuration as in this report~\cite{feature_scattering_defense}. For CIFAR10 and SVHN, we set $(\gamma=1.0,\lambda_1=0.5,\lambda_2=0.7)$. For CIFAR100, we set $(\gamma=1.5,\lambda_1=0.3,\lambda_2=0.42)$. We evaluate the models at the end of the training. The results are summarized in Table~\ref{tab:table-5}.


The joint application of AVmixup with Feature Scatter results in significantly higher accuracy than with Feature Scatter alone. Specifically, on CIFAR10, the combination shows powerful adversarial robustness of 82.31\% and 81.88\% for PGD20 and PGD100, respectively. Furthermore, our experiments on SVHN demonstrate state-of-the-art robustness against the PGD and CW attacks.
Moreover, in contrast with the experimental results of the models trained with the PGD-based approach,
the combination of AVmixup and Feature Scatter shows a significant improvement not only for PGD attacks but also for CW attacks. 

Note that the results on CIFAR100 differ from those on CIFAR10 or SVHN.
The combination also provides state-of-the-art accuracy in all respects, but the increase in accuracy for PGD and CW is small compared to that for other datasets.
We can infer the reason for these results from Table~\ref{table6}, which indicates that the combination is sensitive to the label-smoothing factors.
In this respect, as the number of labels of the dataset increases, the sensitivity of the combination to soft label values increases, which may destabilize the effect of AVmixup.
In addition, we can see that the accuracy on FGSM is slightly higher than that on Clean.
This is because of the property of AVmixup, not because of label leaking~\cite{label_leak}, since the feature scattering-based approach prevents label leaking.
Furthermore, as a larger scaling factor is set for CIFAR100 than for the other datasets that are tested, the average of the vectors used as the training samples is farther away in the adversarial direction from the raw input vector by the larger scaling factor, and this, combined with the property of AVmixup discussed in Section~\ref{discussion1}, can cause such a result. Further discussions of the results can be found in the supplementary material.



\section{Conclusion}

In this work, we identified AFO, the phenomenon that leads to poor robust generalization, and used both theoretical and empirical approaches to show the extent to which soft labeling can help improve robust generalization.
We also introduced AVmixup, a soft-labeled data augmentation method, and demonstrated its outstanding performance through extensive experiments. 
Although AVmixup has shown its excellence in various experiments, the disadvantage of AVmixup is that it imports and uses the linear interpolation without considering the nature of the adversarial examples. 
This forces the appropriate hyperparameters for AVmixup to be found by line search or exhaustive search, and this task will be time consuming if there are limited resources available.
Therefore, we aim to develop advanced algorithms by analyzing in detail the meaning and effects of linear interpolation in AVmixup for future research.

\balance

{\small
\bibliographystyle{ieee_fullname}

}

\newpage
\appendix
\appendixpage
\addappheadtotoc
\newcommand\tab[1][1cm]{\hspace*{#1}}
\newcommand\smalltab[1][1mm]{\hspace*{#1}}

\section{Proofs}
\newtheorem*{theorem1}{Theorem 1}
\begin{theorem1} \label{theorem1_app}
    For the variance parameters $\sigma_r$ and $\sigma_s$, let $\sigma_r=\nu\sigma_s$ where $\nu \in [0,1]$. Then, the upper bound on the standard classification error of $\mathnormal{f}_{n,\sigma_s}$ and the upper bound on the $\ell^{\epsilon}_\infty\text-$robust classification error of $\mathnormal{f}_{n,\sigma_r}$ are equal with probability at least $\left( 1-2\exp(-\frac{d}{8(\sigma_s^2+1)}) \right)\cdot\left( 1-2\exp(-\frac{d}{8(\sigma_r^2+1)}) \right)$ if
    \begin{equation}
        \epsilon \leq \frac{(2\sqrt{n}-1)(1-\nu)}{2\sqrt{n}+4\sigma_s}.
    \end{equation}
\end{theorem1}

\begin{proof}

We recall the following theorems due to Schmidt~\etal~\cite{more_data_adversarial}:

\newtheorem*{theorem18}{Theorem 18}
\begin{theorem18}
    \textup{(Schmidt~\etal)} Let $(\bm{\mathnormal{x}}_1, \mathnormal{y}_1),\dots,(\bm{\mathnormal{x}}_n, \mathnormal{y}_n) \in \mathbb{R}^d\times\{\pm1\}$ be drawn i.i.d. from a $(\theta^{\star{}}, \sigma)\text-$Gaussian model with $\|\theta^\star{}\|_2=\sqrt{d}$. Let the weight vector $\bm{\mathnormal{w}} \in \mathbb{R}^d$ be the unit vector in the direction of $\Bar{\bm{\mathnormal{z}}}=\frac{1}{n}\sum^{n}_{i=1}\mathnormal{y}_i\bm{\mathnormal{x}}_i$.
    Then, with probability at least $1-2\exp(-\frac{d}{8(\sigma^2+1)})$, the linear classifier $f_{\bm{\mathnormal{w}}}$ has classification error at most
    \begin{equation}\label{eq2}
        \exp{\bigg(-\frac{(2\sqrt{n}-1)^2d}{2(2\sqrt{n}+4\sigma)^2\sigma^2}\bigg)}.
    \end{equation}
\end{theorem18} \label{theorem18}

\newtheorem*{theorem21}{Theorem 21}
\begin{theorem21}
\label{theorem21}
    \textup{(Schmidt~\etal)} Let $(\bm{\mathnormal{x}}_1, \mathnormal{y}_1),\dots,(\bm{\mathnormal{x}}_n, \mathnormal{y}_n) \in \mathbb{R}^d\times\{\pm1\}$ be drawn i.i.d. from a $(\theta^{\star{}}, \sigma)\text-$Gaussian model with $\|\theta^\star{}\|_2=\sqrt{d}$. Let the weight vector $\bm{\mathnormal{w}} \in \mathbb{R}^d$ be the unit vector in the direction of $\Bar{\bm{\mathnormal{z}}}=\frac{1}{n}\sum^{n}_{i=1}\mathnormal{y}_i\bm{\mathnormal{x}}_i$.
    Then, with probability at least $1-2\exp(-\frac{d}{8(\sigma^2+1)})$, the linear classifier $f_{\bm{\mathnormal{w}}}$ has $\ell^{\epsilon}_\infty\text-$robust classification error at most $\beta$ if
    \begin{equation}\label{eq3}
        \epsilon \leq \frac{2\sqrt{n}-1}{2\sqrt{n}+4\sigma} - \frac{\sigma\sqrt{2\log{\nicefrac{1}{\beta}}}}{\sqrt{d}}.
    \end{equation}
\end{theorem21}
\noindent
When the upper bound on the standard classification error of $\mathnormal{f}_{n,\sigma_s}$ and the upper bound on the $\ell^{\epsilon}_\infty\text-$robust classification error of $\mathnormal{f}_{n,\sigma_r}$ are equal, 
we can find the condition of $\epsilon$ by substituting (\ref{eq2}) into $\beta$ in (\ref{eq3}).
Then, the right-hand side of (\ref{eq3}) can be written as
    \begin{equation}
    \frac{2\sqrt{n}-1}{2\sqrt{n}+4\sigma_r} - \frac{\sigma_r}{\sigma_s}\cdot\frac{2\sqrt{n}-1}{2\sqrt{n}+4\sigma_s}.
    \end{equation}
Since $\sigma_r=\nu\sigma_s$ where $\nu \in [0,1]$, it satisfies
    \begin{equation}
        \begin{gathered}
        \frac{2\sqrt{n}-1}{2\sqrt{n}+4\sigma_r} -   \frac{\sigma_r}{\sigma_s}\cdot\frac{2\sqrt{n}-1}{2\sqrt{n}+4\sigma_s}
        \geq
        \frac{2\sqrt{n}-1}{2\sqrt{n}+4\sigma_s}\\ -  \frac{\sigma_r}{\sigma_s}\cdot\frac{2\sqrt{n}-1}{2\sqrt{n}+4\sigma_s}
        = 
        \frac{(2\sqrt{n}-1)(1-\nu)}{2\sqrt{n}+4\sigma_s}.
        \end{gathered}
    \end{equation}
\end{proof}


\newtheorem*{corollary1}{Corollary 1}
\begin{corollary1}
    For the variance parameters $\sigma_r$ and $\sigma_s$, let $\sigma_r=\nu\sigma_s$ where $\nu \in [0,1]$. Let the upper bound on the standard classification error of $\mathnormal{f}_{n,\sigma_s}$ and the upper bound on the $\ell^{\epsilon}_\infty\text-$robust classification error of $\mathnormal{f}_{n,\sigma_r}$ be equal. Then, as $\sigma_r$ decreases, the upper bound of $\epsilon$ increases in proportion to $\pi_{n,\sigma_s}$, which is given by
    \begin{equation}
        \pi_{n,\sigma_s}=\frac{2\sqrt{n}-1}{\sigma_s(2\sqrt{n}+4\sigma_s)}.
    \end{equation}
\end{corollary1}

\begin{proof}
Let $\epsilon'$ be the upper bound on $\epsilon$ in Theorem \ref{theorem1_app}. Then, the gradient of $\epsilon'$ with respect to $\sigma_r$ is
    \begin{equation}
    \frac{\partial\epsilon'}{\partial\sigma_r}
    =
    \frac{\partial\epsilon'}{\partial\nu}\frac{\partial\nu}{\partial\sigma_r}
    =-\frac{2\sqrt{n}-1}{2\sqrt{n}+4\sigma_s}\cdot\frac{1}{\sigma_s}
    \end{equation}
Therefore, when $\sigma_r$ decreases, the upper bound of $\epsilon$ increases in proportion to $-\cfrac{\partial\epsilon'}{\partial\sigma_r}$
, which is $\pi_{n,\sigma_s}$.  
\end{proof}

\newtheorem*{theorem2}{Theorem 2}
\begin{theorem2}
\label{theorem2_app}
    Let $\bm{\mathnormal{w}}_B \in \mathbb{R}^{d-c}$ be the weight vector for the sufficient non-robust features of $\Psi_{sample,c}$. Let $Z_{sc}$ be a set of strictly convex functions. Then, when the classifier $\mathnormal{f}_{Z_{sc}}$ is trained on $\Psi_{sample,c}$, the $\bm{\mathnormal{w}}^{\star}_B$ which minimizes the variance of $\bm{\mathnormal{w}}^T\bm{\mathnormal{x}}$ with respect to $\Psi_{sample,c}$ is
    \begin{equation}
        \bm{\mathnormal{w}}^{\star}_B = \Vec{0}.
    \end{equation}
\end{theorem2}

\begin{proof}
Let $\Sigma \in \mathbb{R}^{(d+1) \times (d+1)}$ be the diagonal matrix, where $\Sigma=diag(\sigma_1^2, \cdots, \sigma_{d+1}^2)$ and $\sigma_{i}^2$ is the variance of each $\mathnormal{x}_i$ with $i \in \{1,\cdots,d+1\}$.
Then, the optimization problem is
    \begin{equation} 
    \min_{\bm{\mathnormal{w}}} \smalltab
    \bm{\mathnormal{w}}^T\Sigma\bm{\mathnormal{w}} \quad
    s.t. \smalltab \bm{\mathnormal{w}} \in \mathbb{R}_{+}^{d+1}, \|\bm{\mathnormal{w}}\|_1=1.
    \end{equation}
As the objective function over the feasible set is strictly convex, one optimal weight $\bm{\mathnormal{w}}^\star$ exists at the most.
In addition, we can prove that the optimal weight values for features having the same distribution are equal by using Jensen's Inequality. Thus, we obtain 
    \begin{equation}\label{eq23}
        \begin{gathered}
        \mathnormal{w}_i^\star = \mathnormal{w}_j^\star, \smalltab
        \forall \mathnormal{i}, \mathnormal{j} \in \{1,\cdots,c+1\}, \smalltab or\\ 
        \forall \mathnormal{i}, \mathnormal{j} \in \{c+2,\cdots,d+1\}.
        \end{gathered}
    \end{equation}
We now let $\mathnormal{w}^\star_A$ be the optimal weight value of $\mathnormal{x}_i$, where $\forall\mathnormal{i}\in \{1,\cdots,c+1\}$, and we let $\mathnormal{w}^\star_B$ be the optimal weight value of $\mathnormal{x}_j$, where $\forall\mathnormal{j}\in \{c+2,\cdots,d+1\}$.
Let $\sigma^\star_z$ be the optimal~(minimal) variance of $\bm{\mathnormal{w}}^T\bm{\mathnormal{x}}$. Then, we have ${\sigma^\star_z}^2=(c+1){\mathnormal{w}^\star_A}^2\sigma_A^2+(d-c){\mathnormal{w}^\star_B}^2\sigma_B^2$. 
Thus, ${\sigma^\star_z}^2$ is the same as the solution of the optimization problem
    \begin{equation} 
    \begin{gathered}
    \min_{\mathnormal{w}_B} \smalltab (c+1)\mathnormal{w}_A^2\sigma_A^2+(d-c)\mathnormal{w}_B^2\sigma_B^2\\
         s.t. \smalltab (c+1)\mathnormal{w}_A+(d-c)\mathnormal{w}_B=1.
    \end{gathered}
    \end{equation}
The constraint results in $\mathnormal{w}_A=(1-(d-c)\mathnormal{w}_B) / (c+1)$, which enable us to rewrite the optimization problem as  
    \begin{equation}
    \min_{\mathnormal{w}_B} \smalltab 
    \frac{(1-(d-c)\mathnormal{w}_B)^2}{c+1}\sigma_A^2+(d-c)\mathnormal{w}_B^2\sigma_B^2.
    \end{equation}
Then, we can find the optimal $\mathnormal{w}^{\star}_B$
as follows:   
    \begin{equation}
    \begin{gathered}
    \frac{\mathnormal{d}}{\mathnormal{d}\mathnormal{w}_B}\bigg(\frac{(1-(d-c)\mathnormal{w}_B)^2}{c+1}\sigma_A^2+(d-c)\mathnormal{w}_B^2\sigma_B^2\bigg)  
    \\=2\mathnormal{w}_B\bigg(\frac{(d-c)^2\sigma_A^2}{c+1}+(d-c)\sigma_B^2\bigg)\\-2\bigg(\frac{(d-c)\sigma_A^2}{c+1}\bigg)
    =0.
    \end{gathered}
    \end{equation}
Since $0<\sigma_A\ll\sigma_B$, we have the optimal $\bm{\mathnormal{w}}^{\star}_B$
    \begin{equation}
    \bm{\mathnormal{w}}^{\star}_B=\frac{\sigma_A^2}{(d-c)\sigma_A^2+(c+1)\sigma_B^2}\cdot \Vec{1}
    \approx
    \Vec{0}.
    \end{equation}
as desired.

\end{proof}

\newtheorem*{theorem3}{Theorem 3}
\begin{theorem3}
\label{theorem3_app}
    Let $\bm{\mathnormal{w}}_B \in \mathbb{R}^{d-c}$ be the weight vector for the sufficient non-robust features of $\Psi_{sample,c}$. Let $Z_{sc}$ be a set of strictly convex functions. Then, when the classifier $\mathnormal{f}_{Z_{sc}}$ is trained on $\Psi_{sample,c}$, the $\bm{\mathnormal{w}}^{\star}_B$ which minimizes the variance of $\bm{\mathnormal{w}}^T\bm{\mathnormal{x}}$ with respect to $\Psi_{true}$ is
    \begin{equation}
        \bm{\mathnormal{w}}^{\star}_B = \frac{c}{cd+2c+1} \cdot \Vec{1}.
    \end{equation}
\end{theorem3}

\begin{proof}
As with Theorem \ref{theorem2_app}, let $\sigma_z$ be the variance of $\bm{\mathnormal{w}}^T\bm{\mathnormal{x}}$. Because we train the classifier $\mathnormal{f}_{Z_{sc}}$ on $\Psi_{sample,c}$ but minimize the variance of $\bm{\mathnormal{w}}^T\bm{\mathnormal{x}}$ with respect to $\Psi_{true}$, we have ${\sigma^\star_z}^2={\mathnormal{w}^\star_A}^2\sigma_A^2+c{\mathnormal{w}^\star_A}^2\sigma_B^2+(d-c){\mathnormal{w}^\star_B}^2\sigma_B^2$. Thus, ${\sigma^\star_z}^2$ is the same as the solution of the optimization problem
    \begin{equation} 
    \begin{gathered}
    \min_{\mathnormal{w}_B} \smalltab \mathnormal{w}_A^2\sigma_A^2+c\mathnormal{w}_A^2\sigma_B^2+(d-c)\mathnormal{w}_B^2\sigma_B^2 \\
         s.t. \smalltab (c+1)\mathnormal{w}_A+(d-c)\mathnormal{w}_B=1.
    \end{gathered}
    \end{equation}
The constraint results in $\mathnormal{w}_A=(1-(d-c)\mathnormal{w}_B) / (c+1)$, which enable us to rewrite the objective function as
    \begin{equation} \label{eq17}
    \frac{(1-(d-c)\mathnormal{w}_B)^2}{(c+1)^2}(\sigma_A^2+c\sigma_B^2)+(d-c)\mathnormal{w}_B^2\sigma_B^2.
    \end{equation}
Then, we can find the optimal $\mathnormal{w}^{\star}_B$ as follows:   
    \begin{equation}
    \begin{gathered}
    \frac{\mathnormal{d}}{\mathnormal{d}\mathnormal{w}_B}\bigg(\frac{(1-(d-c)\mathnormal{w}_B)^2}{(c+1)^2}(\sigma_A^2+c\sigma_B^2)+(d-c)\mathnormal{w}_B^2\sigma_B^2\bigg) \\ =
    2\mathnormal{w}_B\bigg(\frac{(d-c)^2}{(c+1)^2}(\sigma_A^2+c\sigma_B^2)+(d-c)\sigma_B^2\bigg)\\-2\bigg(\frac{(d-c)(\sigma_A^2+c\sigma_B^2)}{(c+1)^2}\bigg) = \Vec{0}.
    \end{gathered}
    \end{equation}
Since $0<\sigma_A\ll\sigma_B$, we have the optimal $\bm{\mathnormal{w}}^{\star}_B$
    \begin{equation}
    \begin{gathered}
    \bm{\mathnormal{w}}^{\star}_B=\frac{\sigma_A^2+c\sigma_B^2}{(d-c)(\sigma_A^2+c\sigma_B^2)+(c+1)^2\sigma_B^2} \cdot \Vec{1}
    \\\approx
    \frac{c}{cd+2c+1} \cdot \Vec{1}.
    \end{gathered}
    \end{equation}
as desired.
If we confine our feasible set to 
    \begin{equation} \label{eq33}
    \begin{gathered}
    \mathnormal{w}_i = \mathnormal{w}_j, \smalltab
    \forall \mathnormal{i}, \mathnormal{j} \in \{1,\cdots,c+1\}, \smalltab \\or \smalltab
    \forall \mathnormal{i}, \mathnormal{j} \in \{c+2,\cdots,d+1\},
    \end{gathered}
    \end{equation}
which does not change the optimal weight $\bm{\mathnormal{w}}^\star$ we can easily predict the robust generalization performance in the robust learning procedure using (\ref{eq17}).
\end{proof}

\newtheorem*{lemma1}{Lemma 1}
\begin{lemma1}
\label{lemma1}
    \textup{(Tsipras~\etal)} Minimizing the adversarial empirical risk results in a classifier that assigns 0 weight to non-robust features.
\end{lemma1}

\begin{proof} 
We can follow the proof of Tsipras~\etal~\cite{odds_with_accuracy}, \ie,
let $\mathcal{L}(\bm{\mathnormal{x}},\mathnormal{y};\bm{\mathnormal{w}})$ be the loss function of our classifier $\mathnormal{f}_{\bm{\mathnormal{w}}}$. Then, 
the optimization problem of our model in adversarial training is
    \begin{equation} 
    \min_{\bm{\mathnormal{w}}} \smalltab 
    \mathbb{E}\left[\max_{\|\delta\|_{\infty}\leq\epsilon} \mathcal{L}(\bm{\mathnormal{x}+\delta},y;\bm{\mathnormal{w}})\right].
    \end{equation}
Consider any weight $\bm{\mathnormal{w}}$ for which $\mathnormal{w}_i > 0$ for some $i \geq c+2$. Because the sufficient non-robust features $\mathnormal{x}_{c+2},\dots,\mathnormal{x}_{d+1}$ are normally distributed random variables with mean $\eta\mathnormal{y}$,
the adversary reverses the sign of the sufficient non-robust features if $\eta < \epsilon$.
Then, it assigns negative update values with a high probability, because the sufficient non-robust features are moved such that their signs are opposite to that of the label.
Thus, the optimizer constantly updates the weight vector such that the weight values $\mathnormal{w}_i$, where $\forall i \geq c+2$, move toward 0. Ultimately, the optimization of the objective function assigns 0 weight to the non-robust features.

\end{proof}

\section{Dataset}

\begin{table*}[h]
\centering
\caption{Dataset configuration used in the experiments.}
\begin{tabular}{cccccc}
\toprule
\multicolumn{1}{c|}{Dataset} & Train Size & Test Size & Class Size & Image Size \\ \hline
\multicolumn{1}{c|}{CIFAR10} & 50,000 & 10,000  & 10   & (32, 32) \\
\multicolumn{1}{c|}{CIFAR100}& 50,000 & 10,000 & 100 & (32, 32) \\
\multicolumn{1}{c|}{SVHN}    & 73,257  & 26,032 & 10 & (32, 32) \\
\multicolumn{1}{c|}{Tiny Imagenet}    & 100,000 & 20,000 & 200 & (64, 64) \\
\bottomrule
\end{tabular}

\label{tab:dataset}
\end{table*}

\begin{table*}[!htb]
\centering
\caption{Accuracy comparisons of our proposed approach AVmixup with PGD~\cite{pgd_attack} and LS$\lambda$~($\lambda \in \{0.8,0.9\}$)~\cite{inception} against white-box attacks on CIFAR100.}
\begin{tabular}{cccccc}
\toprule
\multicolumn{1}{c|}{Model} & Clean & FGSM & PGD10 & PGD20 & CW20 \\ \hline
\multicolumn{1}{c|}{Standard} & 78.57 & 6.56  & 0.0   & 0.0   & 0.0 \\
\multicolumn{1}{c|}{PGD}      & 61.29 & 46.04 & 25.76 & 25.17 & 22.98 \\
\multicolumn{1}{c|}{LS0.8}    & 62.1  & 52.33 & 29.47 & 28.81 & \textbf{26.15} \\
\multicolumn{1}{c|}{LS0.9}    & 61.77 & 53.17 & 27.71 & 27.13 & 25.34 \\
\multicolumn{1}{c|}{AVmixup} & \textbf{74.81}       & \textbf{62.76}                      & \textbf{39.98}       & \textbf{38.49}       & 23.46       \\ \bottomrule
\end{tabular}
\label{tab:whitebox-cifar100}
\end{table*}

We conducted experiments on the CIFAR10~\cite{cifar_dataset}, CIFAR100~\cite{cifar_dataset}, SVHN~\cite{svhn_dataset}, and Tiny Imagenet~\cite{imagenet_dataset} datasets. As shown in Table~\ref{tab:dataset}, the CIFAR10 and SVHN datasets comprise fewer classes and more data per class than other datasets such as the CIFAR100 and Tiny Imagenet datasets. Thus, the classification task of CIFAR100 and Tiny Imagenet are significantly more difficult than the tasks of CIFAR10 and SVHN.  

\section{Further Related Work}

In this section, we introduce two recent works~\cite{iat, mixup_infer} that utilized Mixup~\cite{mixup} in adversarial defense. Pang~\etal~\cite{mixup_infer} developed Mixup Inference~(MI), an inference principle, by focusing on the globally linear behavior in-between the data manifolds introduced by the mixup training method. MI feeds the linear interpolation between the adversarial example and a sampled clean example into the classifier. The main difference between AVmixup and MI is that MI is an algorithm applied only in the inference step while AVmixup is applied only in the training step.

Lamb~\etal~\cite{iat} proposed Interpolated Adversarial Training~(IAT), the interpolation based adversarial training. IAT takes two losses in a training step, one is from interpolations between clean examples, and the other is from that between adversarial exampels. AVmixup, on the other hand, takes interpolations between the clean example and its adversarial vertex. Ultimately, IAT improves standard test error while maintaining the robustness, and AVmixup simultaneously improves both the standard accuracy and the adversarial robustness.

\section{Detailed Experimental Results}
\label{sec:detailed_results}

We present the detailed results of accuracy against white-box attacks on CIFAR100 in Table~\ref{tab:whitebox-cifar100}. AVmixup significantly improves the accuracy for Clean, FGSM, and PGD, but its improvement on CW20 is limited.

Additionally, we also provide the results of accuracy against black-box attacks on CIFAR10 and CIFAR100 in Tables~\ref{tab:blackbox-cifar10} and~\ref{tab:blackbox-cifar100}, respectively. The models of each column represent the attack models of the transfer-based black-box attacks, and the models of each row represent the evaluated defense models. It is evident that the AVmixup model is the most robust against black-box attacks from all of the attack models with significant margins.

\begin{table*}[h]
\centering
\caption{Accuracy comparisons against transfer-based black-box attacks on CIFAR10.}
\begin{tabular*}{\textwidth}{c @{\extracolsep{\fill}} ccccccccccc}
\toprule
\multicolumn{1}{l}{}        & \multicolumn{5}{c}{PGD20}                                                                 & \multicolumn{5}{c}{CW20}         \\ \cline{2-11} 
Model                      & Standard & PGD            & LS0.8          & LS0.9         & AVmixup                              & Standard & PGD & LS0.8 & LS0.9 & AVmixup \\ \hline
\multicolumn{1}{c|}{PGD}    & 85.6     & -          & 65.70           & 64.91         & \multicolumn{1}{l|}{72.30}           & 85.77      & - & 66.32   & 65.68   & 72.25    \\
\multicolumn{1}{c|}{LS0.8}  & 86.03    & 63.6           & -          & 64.83         & \multicolumn{1}{l|}{72.34}          & 86.02      & 64.37 & -   & 65.51   & 72.28    \\
\multicolumn{1}{c|}{LS0.9}  & 86.4     & 63.74          & 65.78          & -         & \multicolumn{1}{l|}{72.67} & 86.59      & 64.46 & 66.31   & -   & 72.82    \\
\multicolumn{1}{c|}{AVmixup} & \textbf{89.53}    & \textbf{68.51} & \textbf{71.48} & \textbf{70.50} & \multicolumn{1}{l|}{\quad-}          & \textbf{89.65}      & \textbf{69.40} & \textbf{71.72}   & \textbf{70.97}   & -    \\ \bottomrule
\end{tabular*}
\label{tab:blackbox-cifar10}
\end{table*}

\begin{table*}[h]
\centering
\caption{Accuracy comparisons against transfer-based black-box attacks on CIFAR100.}
\begin{tabular*}{\textwidth}{c @{\extracolsep{\fill}} ccccccccccc}
\toprule
\multicolumn{1}{l}{}        & \multicolumn{5}{c}{PGD20}                                                                 & \multicolumn{5}{c}{CW20}         \\ \cline{2-11} 
Model                      & Standard & PGD            & LS0.8          & LS0.9         & AVmixup                              & Standard & PGD & LS0.8 & LS0.9 & AVmixup \\ \hline
\multicolumn{1}{c|}{PGD}    & 51.59     & -          & 36.68           & 36.74         & \multicolumn{1}{l|}{40.88}       & 51.97      & - & 37.99   & 38.62   & 40.15    \\
\multicolumn{1}{c|}{LS0.8}  & 60.27    & 41.44           & -          & 37.61         & \multicolumn{1}{l|}{46.27}          & 60.35      & 42.15 & -   & 38.01   & 44.6    \\
\multicolumn{1}{c|}{LS0.9}  & 59.93 & 40.68          & 36.86          & -         & \multicolumn{1}{l|}{45.73} & 60.3      & 41.61 & 36.97   & -   & 44.42    \\
\multicolumn{1}{c|}{AVmixup} & \textbf{67.69}    & \textbf{44.78} & \textbf{44.86} & \textbf{45.3} & \multicolumn{1}{l|}{\quad-}          & \textbf{68.02}      & \textbf{43.83} & \textbf{44.8}   & \textbf{45.95}   & -    \\ \bottomrule
\end{tabular*}
\label{tab:blackbox-cifar100}
\end{table*}

\section{Empirical Evidence on the Nature of Mixup}
\label{sec:emp_mixup}

We previously mentioned that utilizing virtual data constructed by linear interpolation promotes the tight generalization of features observed in the training steps.
In addition, we demonstrated that the accuracy of the combination of AVmixup with a feature scattering-based approach~(Feature Scatter) against FGSM on CIFAR100 is slightly higher than that on Clean.
In this section, we introduce the experimental results, which support the hypothesis.
The following settings are compared in the experiment:
\begin{enumerate}[leftmargin=*, nolistsep]
    \item Standard: The model that is trained with the original dataset. 
    \item Mixup: With the original dataset, we apply Mixup~\cite{mixup} for the model.
    \item Gvrm: The model that is trained by Vicinal Risk Minimization~(VRM) with a Gaussian kernel~\cite{vrm}.
    \item Noisy mixup: With the Gaussian-noise-added dataset, we apply Mixup~\cite{mixup} for the model.
\end{enumerate}
The original test set and the Gaussian-noise-added-test set are denoted as Clean and Noise, respectively, and we used the same Gaussian distribution $\mathnormal{N}(0, 8^2)$ for both the training and evaluation procedures.

Based on Table~\ref{tab:mixup}, we confirm the strong generalization performance of the mixup algorithm. In other words, the Mixup model and the Noisy mixup model show the highest accuracy on Clean and Noise, respectively.

To understand the property of AVmixup, we should focus on the results of the Noisy mixup model, which shows the worst accuracy on Clean and the best accuracy on Noise.
Moreover, the accuracy on Noise is significantly higher than that on Clean by 2.29\%p.
This implies that the Noisy mixup model tightly generalizes the input distribution, which is slightly changed owing to Gaussian noise during the training.
Therefore, AVmixup with the PGD-based approach does not show a high level of robustness against other types of adversarial attacks, and the accuracy of AVmixup with Feature Scatter against FGSM on CIFAR100 can be higher than that on Clean for a high scaling factor~($\gamma=1.5$).

\begin{table}[h]
\centering
\caption{Accuracy~(median over 5 runs) comparisons on Clean and Noise.}
\begin{tabular}{ccccc}
\toprule
\multicolumn{1}{c|}{Model}  & Clean        & Noise    \\ \hline
\multicolumn{1}{c|}{Standard}  & 94.52         & 85.8                \\
\multicolumn{1}{c|}{Mixup}  & \textbf{95.93}        & 88.97     \\
\multicolumn{1}{c|}{Gvrm}  & 92.77        & 92.96             \\
\multicolumn{1}{c|}{Noisy mixup} & 92.56        & \textbf{94.85}              \\
 \bottomrule
\end{tabular}
\label{tab:mixup}
\end{table}

\begin{table}[h]
\centering
\caption{Accuracy results of AVmixup with PGD-based approach on CIFAR10. $\gamma$ is the scaling factor of AVmixup, and $\lambda_1$ and $\lambda_2$ are label-smoothing factors for the raw input and the adversarial vertex, respectively.}
\begin{tabular}{ccc|ll}
\toprule 
\multicolumn{1}{c|}{}           & \multicolumn{2}{c|}{$\gamma=1.0$}                                        & \multicolumn{2}{c}{$\gamma=2.0$}         \\ 
\multicolumn{1}{c|}{$\lambda_1$ / $\lambda_2$}          & Clean                     & PGD10                               & Clean          & PGD10          \\ \hline
\multicolumn{1}{c|}{1.0 / 0.1} & \textbf{92.88}            & 7.43                                & \textbf{92.63} & \textbf{53.19} \\
\multicolumn{1}{c|}{1.0 / 0.5} & 91.29                     & 42.49                               & 92.04          & 49.33          \\
\multicolumn{1}{c|}{0.5 / 0.1} & 92.59                     & 14.07                               & 91.78          & 51.43          \\
\multicolumn{1}{c|}{0.5 / 0.7} & \multicolumn{1}{c}{90.12} & \multicolumn{1}{c|}{\textbf{47.36}} & 84.93          & 53.06         
\\ \bottomrule
\end{tabular}
\label{tab:pgd}
\end{table}

\section{Ablation Study on CIFAR10}
\label{sec:ablation_study}


To analyze the sensitivity of AVmixup to the hyperparameters, we conducted an ablation study, and the results are summarized in Tables~\ref{tab:pgd} and \ref{tab:feature-scatter}. As shown in both tables, the larger the $\lambda_1$, the higher is the accuracy on Clean. Furthermore, $\lambda_1$ and $\lambda_2$ imply the weights on standard accuracy and adversarial robustness, respectively. However, they do not fit perfectly as changes occured by other conditions.


From the comparisons based on $\gamma$, it is evident that AVmixup with the PGD-based and feature scattering-based approaches yielded the highest accuracy at $\gamma=1$ and at $\gamma=2$, respectively. This is owing to the difference in the construction of adversarial examples between the two approaches. Because the PGD-based approach constructs adversarial examples using a multi-step method, adversarial examples are located in the $\ell_\infty\text-$bounded cube. Otherwise, because the feature scattering-based approach constructs adversarial examples using a one-step method, adversarial examples are located on the surface of the $\ell_\infty\text-$bounded cube. Therefore, with the PGD-based approach, AVmixup requires a larger scaling factor to cover the $\ell_\infty\text-$bounded space. Likewise, with the feature scatter-based approach, AVmixup requires the scaling factor of 1 to cover the $\ell_\infty\text-$bounded space.

The sensitivity of AVmixup is largely related to the label-smoothing factors, which tune the input data labels for robust generalization based on our theoretical results. They are also connected to the Lipschitz constant for the classifier, with respect to the training samples in adversarial directions. This complex relationship causes the observed high-sensitivity. This may be interpreted as a fault of AVmixup, but also provides meaningful insight into useful directions for future research.

\begin{table*}[h]
\centering
\caption{Accuracy results of AVmixup with feature scattering-based approach on CIFAR10.}

\begin{tabular}{ccccc|llll}
\toprule
\multicolumn{1}{c|}{}           & \multicolumn{4}{c|}{$\gamma=1.0$}                                                                                                & \multicolumn{4}{c}{$\gamma=2.0$}                                           \\ 
\multicolumn{1}{c|}{$\lambda_1$ / $\lambda_2$}          & Clean                              & FGSM                      & PGD20                     & CW20                       & Clean          & FGSM           & PGD20          & CW20           \\ \hline
\multicolumn{1}{c|}{0.1 / 0.5} & 91.94                              & 80.09                     & 74.43                     & 62.87                      & 90.87          & 93.42          & \textbf{72.94} & \textbf{70.11} \\
\multicolumn{1}{c|}{0.5 / 0.3} & 92.82                              & 77.81                     & 57.67                     & 55.18                      & 95.41          & 95.32          & 60.95          & 47.32          \\
\multicolumn{1}{c|}{0.5 / 0.7} & 92.37                              & \textbf{83.49}            & \textbf{82.31}            & \textbf{71.88}             & 95.2           & \textbf{95.86} & 52.87          & 47.79          \\
\multicolumn{1}{c|}{1.0 / 0.5} & \multicolumn{1}{c}{\textbf{93.07}} & \multicolumn{1}{c}{79.55} & \multicolumn{1}{c}{53.42} & \multicolumn{1}{c|}{56.72} & \textbf{95.43} & 95.12          & 47.41          & 44.14        
\\ \bottomrule
\end{tabular}

\label{tab:feature-scatter}
\end{table*}

\section{Code}
\label{sec:code}

Our codes are available at \url{https://github.com/Saehyung-Lee/cifar10_challenge}.

\end{document}